%% file: main.tex
\documentclass{article} 
\usepackage[dvipsnames]{xcolor}

\input{math_commands.tex}

\usepackage{iclr2023_conference,times}

\usepackage[utf8]{inputenc} 
\usepackage[T1]{fontenc}    
\usepackage{hyperref}       
\usepackage{url}            
\usepackage{booktabs}       
\usepackage{amsfonts}       
\usepackage{nicefrac}       
\usepackage{microtype}      
\usepackage{graphicx}
\usepackage{amsmath,amssymb} 
\usepackage{color}
\usepackage{epsfig}
\usepackage{tabularx}
\usepackage{booktabs}
\usepackage{multirow}
\usepackage{array}
\usepackage{xspace}
\usepackage{balance}
\usepackage{enumitem}
\usepackage{mathtools}
\usepackage{rotating}
\usepackage{tabulary}
\usepackage[export]{adjustbox}
\usepackage{ctable}
\usepackage{diagbox}
\usepackage{subcaption}
\usepackage{makecell}
\usepackage[linesnumbered,ruled]{algorithm2e}
\usepackage{bbm}
\usepackage{wrapfig}
\usepackage[title]{appendix}

\newcommand{\ie}{\textit{i}.\textit{e}.}
\newcommand{\eg}{\textit{e}.\textit{g}.}

\graphicspath{{./image/}}

\title{Masked Unsupervised Self-training for Label-free Image Classification 
} 

\author{Junnan Li, Silvio Savarese, Steven Hoi \\
	Salesforce Research\\
	\texttt{\{junnan.li,ssavarese,shoi\}@salesforce.com}
}   

\newcommand{\name}{MUST\xspace}
\iclrfinalcopy

\begin{document}

\maketitle
\vspace{-2ex}

\input{sec_abstract}
\vspace{-1ex}
\input{sec_introduction}
\input{sec_literature}

\input{sec_method}

\input{sec_experiment}

\input{sec_conclusion}

\bibliographystyle{iclr2023_conference}
\bibliography{main.bib}

\end{document}

%% file: math_commands.tex
\usepackage{amsmath,amsfonts,bm}

\def\eqref#1{equation~\ref{#1}}

\def\1{\bm{1}}

\DeclareMathAlphabet{\mathsfit}{\encodingdefault}{\sfdefault}{m}{sl}
\SetMathAlphabet{\mathsfit}{bold}{\encodingdefault}{\sfdefault}{bx}{n}

\DeclareMathOperator*{\argmax}{arg\,max}

%% file: sec_abstract.tex
\begin{abstract}
State-of-the-art computer vision models are mostly trained with supervised learning using human-labeled images,
which limits their scalability due to the expensive annotation cost.
While self-supervised representation learning has achieved impressive progress,
it still requires a second stage of finetuning on labeled data.
On the other hand,
models pre-trained with large-scale text-image supervision (e.g., CLIP) have enabled zero-shot transfer to downstream image classification tasks.
However,
the zero-shot performance of CLIP-like models are often insufficient for real-world adoption.
In this paper,
we aim to leverage the abundant unlabeled data from a target domain to improve the performance of a pre-trained zero-shot classifier, by unsupervised finetuning of the pre-trained model.
We propose Masked Unsupervised Self-Training (\name),
a new unsupervised adaptation method which leverages two different and complementary sources of training signals: pseudo-labels and raw images.
\name jointly optimizes three objectives to learn both class-level global feature and pixel-level local feature and enforces a regularization between the two.
We demonstrate the efficacy of \name on a variety of downstream tasks,
where it improves upon CLIP by a large margin.
\name also outperforms supervised few-shot adaptation methods.
It achieves a top-1 accuracy of 77.7\% on ImageNet using ViT-B, +9.4\% higher than CLIP, and +6.2\% higher than 16-shot CLIP adaptation.
Our code is available at \url{https://github.com/salesforce/MUST}.
\end{abstract}
	

%% file: sec_introduction.tex
\vspace{-1ex}
\section{Introduction}
\label{sec:introduction}
\vspace{-1ex}
Zero-shot image classification is a challenging goal that marks the capability of a vision model to solve tasks without human supervision.
Recently, 
vision-language pre-training (e.g. CLIP~\citep{clip}) has shown promising performance on open-vocabulary zero-shot classification,
where it leverages web-scale image-text pairs to train image and text encoders that can be transferred to downstream tasks through natural language prompting.
However,
the zero-shot performance of CLIP is often inadequate for real-world adoptions, especially when compared to models that are trained with supervised learning.
On the other hand, there are abundant unlabeled data available for many tasks.
In this paper,
we aim to improve the performance of an open-vocabulary zero-shot classifier by finetuning it on unlabeled images from a downstream task.

Given unlabeled data, the key question is: \textit{what is the source of supervision?}
Numerous papers have attempted to answer this question.
Among them,
\textit{self-training} and \textit{self-supervised learning} are two of the most dominant approaches.
In self-training~\citep{rethink_selftraining,noisy_student,pseudo_label},
pseudo-labels are generated by a teacher model and then used to supervise task-specific training of a student model, 
where the student usually has an equal model size as the teacher.
On the other hand, self-supervised learning methods are generally task-agnostic.
Masked image modeling~\citep{beit,mae,simmim}, which trains the model to predict missing information from masked image patches, has recently emerged as the superior self-supervised learning method for vision transformers (ViT)~\citep{vit}.
However, both self-training and self-supervised learning have their limitations.
Self-training overly relies on the pseudo-labels as the only source of supervision,
thus is prone to overfitting to the noise in pseudo-labels.
Self-supervised learning requires an additional stage of task-specific finetuning on labeled data, thus is not a one-stop solution.

In this paper,
we propose Masked Unsupervised Self-Training (\name),
a simple and effective method for label-free image classification.
\name performs unsupervised learning using both pseudo-labels and raw images as two different and complementary training signals.
Specifically,
\name jointly optimizes three objectives to finetune a pre-trained classification model (\eg~CLIP) on unlabeled images:
(1) Self-training objective to learn global task-specific class prediction;
(2) Masked image modeling objective to learn local pixel-level information;
(3) Global-local feature alignment objective to bridge the knowledge learned from the two sources of supervision.

We validate the efficacy of \name on 8 image classification tasks across a variety of domains,
showing significant improvement over CLIP~\citep{clip}.
\name also outperforms supervised few-shot adaptation methods~\citep{coop,cocoop}.
For instance, \name achieves 77.7\% top-1 accuracy on ImageNet, +9.4\% higher than CLIP, and +6.2\% higher than 16-shot CLIP adaptation.
On certain domains, \name can achieve comparable performance with a fully-supervised method.
We further perform extensive quantitative and qualitative analysis to examine the effect of each proposed component.
\name is a low-cost solution for image classification that unlocks the potential of CLIP-like models for practical scenarios where images are abundant but labels are scarce.





%% file: sec_literature.tex
\vspace{-0.5ex}
\section{Related Work}
\label{sec:literature}
\vspace{-1ex}

\noindent\textbf{Zero-shot learning} traditionally aims to recognize unseen classes by training the model on base classes~\citep{zeroshot_benchmark,zeroshot_survey},
where the most common approach is to utilize auxiliary information such as attributes~\citep{zeroshot_attribute} or knowledge graphs~\citep{zeroshot_kg}.
CLIP~\citep{clip} popularizes a new approach for open-vocabulary zero-shot image classification by leveraging natural language supervision from web-scale datasets.
Despite its impressive performance,
the zero-shot accuracy of CLIP stills falls far below the supervised method in many domains.
Some recent work tries to adapt CLIP to downstream tasks using labeled data~\citep{coop,clip_adapter}, which is less scalable compared to our unsupervised adaptation method.
Our method is also orthogonal to CLIP-like research in model pre-training and can be applied to other image classification models~\citep{align,declip,filip,mopro},
such as LiT~\citep{LiT} which performs two-stage pre-training.

\noindent\textbf{Self-training} has shown promising progress in many domains including vision~\citep{rethink_selftraining,noisy_student,selfself},
NLP~\citep{selftrain_nlp}, and speech~\citep{selftrain_speech}.
Our method is more closely related to the self-training approaches proposed for semi-supervised learning~\citep{mean_teacher,fixmatch,remixmatch},
where pseudo-labels on unlabeled data are used as training targets.
We construct our self-training objective by following three principles:
(1) \textit{consistency regularization}~\citep{fixmatch,semi_temporal} which enforces the model to output the same prediction when the input is perturbed;
(2) \textit{entropy minimization}~\citep{EntMin} which encourages the model to give ``sharp'' predictions with low entropy;
(3) \textit{prediction fairness}~\citep{remixmatch} which alleviates the model's bias towards certain classes.
Our method leverages a separate supervision signal from raw image pixels to reduce the confirmation bias in self-training, which is orthogonal to methods for pseudo-label debiasing~\citep{crest,debias}.

\noindent\textbf{Masked image modeling},
fueled by the success of vision transformers~\citep{vit},
recently emerged as a more appealing self-supervised representation learning method over contrastive learning~\citep{moco,simclr,DIM}.
While some methods train the model to predict discrete tokens~\citep{beit} or contextualized representations~\citep{data2vec} for masked image patches,
MAE~\citep{mae} and SimMIM~\citep{simmim} achieve competitive performance by simply predicting the pixel values. 
The MIM objective has also been used for test-time training~~\citep{ttt}.
Different from existing self-supervised learning methods which require an additional stage of supervised finetuning on labeled data,
we synergically incorporate masked image modeling into self-training as a one-stage solution for zero-shot image classification.


%% file: sec_method.tex
\vspace{-0.5ex}
\section{Method}
\label{sec:method}
\vspace{-1ex}

\input{figure/clip.tex}

\name is a simple unsupervised learning approach that adapts a pre-trained open-vocabulary classifier to a downstream task using unlabeled images.
In this paper,
we consider vision transformers pre-trained by CLIP~\citep{clip} as the models to be adapted due to their distinctive zero-shot performance.
CLIP pre-trains an image encoder and a text encoder with a contrastive loss such that paired images and texts have higher similarities in a shared embedding space.
In order to perform zero-shot classification,
CLIP converts a set of class names into text embeddings using ensemble of natural language prompts (\eg, a photo of a \{object\}).
During inference, it takes the dot-product between an image embedding and all text embeddings to produce the prediction logits for that image.
As shown in Figure~\ref{fig:clip}, 
we convert CLIP's non-parametric text embeddings into weights of a linear classifier, and directly finetune the linear classifier together with the image encoder for unsupervised adaptation. 

Figure~\ref{fig:framework} shows the overall framework of \name.
Given an image, we follow ViT~\citep{vit} to divide it into regular non-overlapping patches.
A \texttt{[CLS]} token is appended to extract global information,
which is used by the classifier for prediction.
We then randomly mask image patches by replacing a patch's embedding with a learnable \texttt{[MSK]} token.
The output embeddings of \texttt{[CLS]} and \texttt{[MSK]} are used to jointly optimize three objectives:
(1) global self-training,
(2) local masked image modeling,
and (3) global-local feature alignment.
Next we delineate the details of each objective.

\subsection{Self-training with an EMA Teacher}
\label{sec:self_train}

The self-training objective is applied to the classifier's output.
Given a batch of $B$ unlabeled images,
we compute a pseudo-label for each image by passing a weakly-augmented version of the image to a teacher model.
The teacher is parameterized by an exponentially moving average (EMA)~\citep{mean_teacher,byol} of the model parameters $\theta$.
Specifically,
the parameters of the EMA teacher $\Delta$ are intialized as $\theta$.
During each update of $\theta$, $\Delta$ are updated with
\begin{equation}
\Delta = \mu \Delta + (1-\mu) \theta
\end{equation}

Let $q_b$ denote the EMA teacher's softmax prediction for the weakly-augmented image,
we enforce a cross-entropy loss against the model's prediction $p_b$ for a strongly-augmented version of the same image:
\begin{equation}
    \mathcal{L}_\mathrm{cls} = \frac{1}{B} \sum_{b=1}^{B} \mathbbm{1} (\max q_b\geq \tau) \mathrm{H}(\hat{q}_b,p_b)
\end{equation}
Following FixMatch~\citep{fixmatch},
we only use pseudo-labels with maximum scores above a threshold $\tau$,
and convert the soft labels $q_b$ into ``one-hot'' hard labels by $\hat{q}_b = \argmax(q_b)$.

Since the pseudo-labels generated by a CLIP model are often biased towards certain classes~\citep{debias}, minimizing $\mathcal{L}_\mathrm{cls}$ alone would magnify the bias.
To mitigate confirmation bias, we introduce a ``fairness'' regularization which encourages that on average, across a batch of samples,
the model's prediction probability is close to a uniform distribution:
\begin{equation}
    \mathcal{L}_\mathrm{reg} = -\frac{1}{K} \sum_{k=1}^K\log(\bar{p}_k),
\end{equation}    
where $K$ is the total number of classes, and $\bar{p}$ is the model's average prediction across the batch.
In cases where $K>B$, we compute $\bar{p}$ using moving average instead of batch average.
We find this regularization to be beneficial even for datasets with long-tailed class distribution (see Table~\ref{tbl:fairness}).

\input{figure/framework}

\subsection{Masked Image Modeling}

To alleviate the over-reliance on the noisy pseudo-labels in self-training,
we introduce another source of supervision obtained from the raw images.
The masked image modeling (MIM) objective aims to learn local image representation at masked positions by predicting the missing information using contextual patches.
We follow SimMIM~\citep{simmim} and simply predict the RGB pixel values for masked patches.
Specifically,
given the output embedding $z_b^m$ of the $m$-th \texttt{[MSK]} token,
we first pass it through a linear decoder head to obtain the predicted RGB values $y_b^m\in \mathbb{R}^N$ for that patch, where $N$ denotes the number of RGB pixels per patch.
Then we compute the MIM loss as an $\ell_1$-loss between $y_b^m$ and the ground-truth RGB values $x_b^m$:
\begin{equation}
    \mathcal{L}_\mathrm{mim} = \frac{1}{BMN} \sum_{b=1}^{B}  \sum_{m=1}^{M} \lVert y_b^m - x_b^m \rVert_1,
\end{equation}
where $M$ denotes the number of masked patches per image. 
We employ the patch-aligned random masking strategy~\citep{simmim} where multiple $s\times s$ patches are randomly masked with a fix masking ratio for each dataset.
In general,
we find that a low masking ratio (e.g., 10\%) works well for most datasets,
which is consistent to the observations in~\citet{simmim}.


\subsection{Global-local Feature Alignment}

We aim to bridge the two sources of supervision (\ie,~pseudo-labels and image pixels) such that the local \texttt{[MSK]} features learned from MIM can improve the global \texttt{[CLS]} feature for better classification performance.
Let $z_b^c$ denote the output embedding of the \texttt{[CLS]} token,
and $v_b^c=h(z_b^c)$ denote its normalized embedding after the projection network $h$.
Following CLIP, $h$ is a linear layer followed by $\ell_2$ normalization,
and $v_b^c$ is subsequently used by the classifier to produce the prediction $p_b$ for self-training (see Section~\ref{sec:self_train}).
We also project the embeddings of the \texttt{[MSK]} tokens to the \textit{same} space by passing them through $h$: $v_b^m=h(z_b^m)$.
The global-local feature alignment loss is defined as the average squared distance between the normalize embeddings of the \texttt{[CLS]} token and all \texttt{[MSK]} tokens for each image: 
\begin{equation}
\label{eqn:align}
    \mathcal{L}_\mathrm{align} = \frac{1}{BM} \sum_{b=1}^{B}  \sum_{m=1}^{M} \lVert v_b^c - v_b^m \rVert_2^2
\end{equation}

During training, \name jointly optimizes the above three objectives. The overall loss is
\begin{equation}
\mathcal{L} = \mathcal{L}_\mathrm{cls} + \lambda_\mathrm{reg}\mathcal{L}_\mathrm{reg} + \mathcal{L}_\mathrm{mim} + \lambda_\mathrm{align} \mathcal{L}_\mathrm{align}
\end{equation}

\subsection{Implementation Details}
\label{sec:implementation}

We experiment with two ViT models pre-trained by~\citet{clip}: ViT-B/16 and ViT-L/14, containing respectively 12 and 24 Transformer blocks with 768 and 1024 hidden dimensions.
The \texttt{[MSK]} token and linear decoder head are randomly initialized and finetuned together with the entire model.  
During finetuning,
we use AdamW~\citep{adamw} optimizer with a weight decay of 0.05.
We employ a cosine learning rate schedule without any warmup.
Following~\citet{beit,mae},
we use a layer-wise learning rate decay~\citep{electra} of 0.65 for both ViT models. 
The batch size is 1024 for ViT-B/16 and 512 for ViT-L/14,
and the learning rate is scaled linearly with the batch size ($lr=\text{base\_lr} \times \text{batchsize} / 256$).
We use 16 A100 GPUs,
and the training process of \name adds only a small amount of computation overhead after CLIP pre-training.
We observe negligible variance on the results between runs with different random seeds.

The model receives input images of size $224\times 224$.
During training,
we use RandomResizedCrop+Flip+RandAug~\citep{randaug} to augment input images,
while using Resize+RandomCrop as the weak augmentation to generate pseudo-labels.
During test,
we simply take a center crop after resizing the shorter edge of the image to 224.
For the EMA teacher, we follow~\citet{data2vec} and linearly ramp-up the parameter decay rate $\mu$ from $\mu_0$ to 0.9998 in $\mu_n$ iterations.
Table~\ref{tbl:hyper_parameters} provides more details of the hyperparameters used for each downstream task, including the pseudo-label threshold, mask patch size, mask ratio, etc.

%% file: figure/clip.tex
\begin{figure*}[!t]
\centering
  \includegraphics[width=\textwidth]{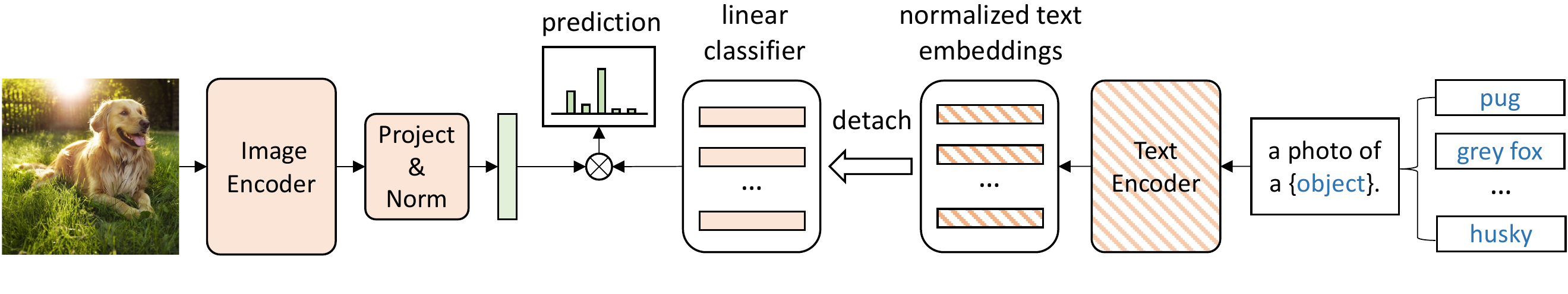}
\vspace{-4ex}
\caption{
\small 
On each downstream task, we convert CLIP's text embeddings into a linear classifier, which is fine-tuned together with the image encoder for unsupervised adaptation.
}
\vspace{-1.5ex}
\label{fig:clip}
\end{figure*}

%% file: figure/framework.tex
\begin{figure*}[!t]
\centering
  \includegraphics[width=\textwidth]{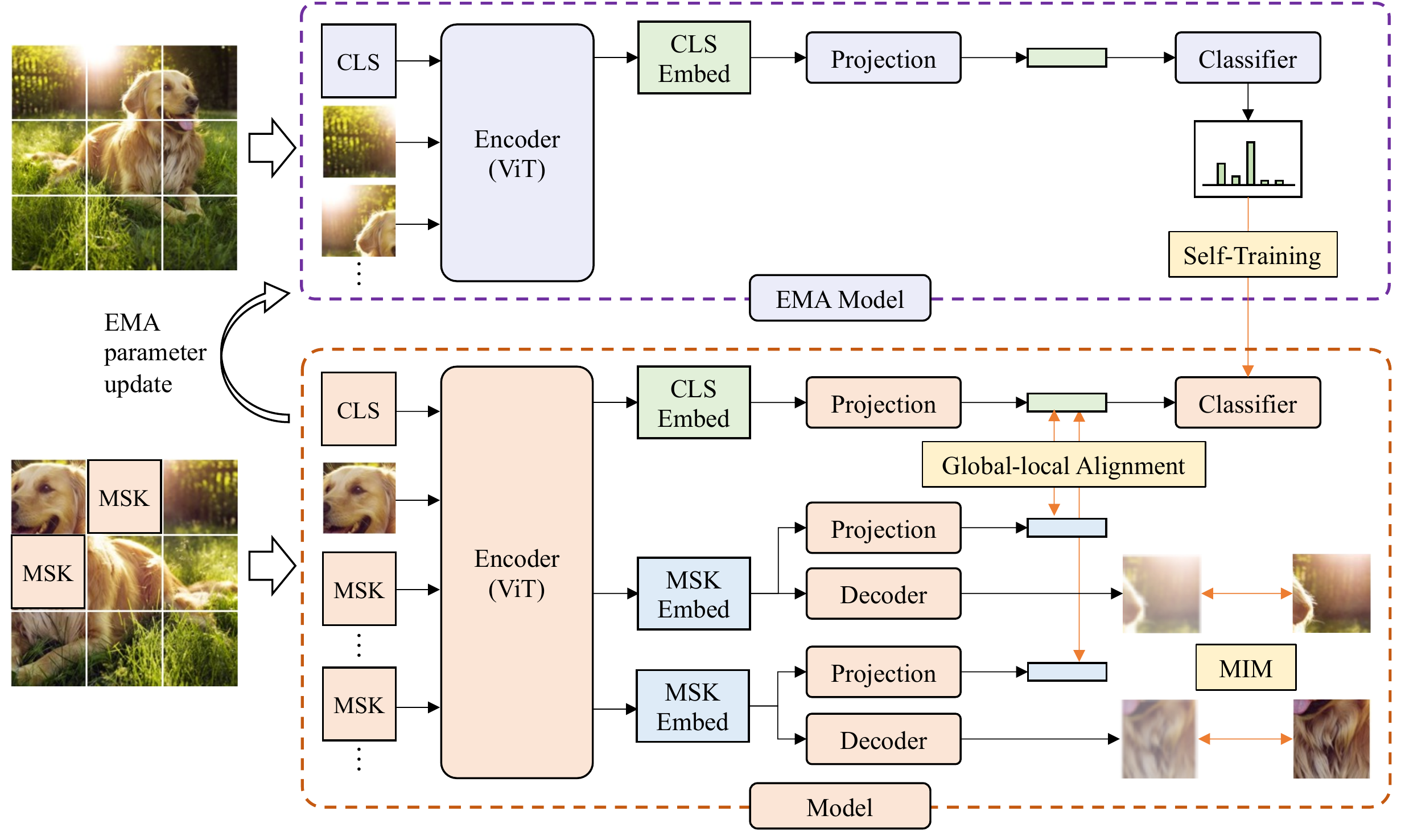}
\vspace{-2ex}
\caption{
\small
Unsupervised learning framework of \name. 
We randomly replace image patches with \texttt{[MSK]} tokens, and
jointly optimize three objectives on unlabeled images.
(1) Global self-training: the classifier uses the output of \texttt{[CLS]} to predict the pseudo-label produced by an EMA model.
(2) Masked image modeling: the linear decoder uses the outputs of \texttt{[MSK]} to predict the pixel values of the masked patches.
(3) Global-local feature alignment: minimize the distance between \texttt{[CLS]} and \texttt{[MSK]} in a normalized embedding space.
}
\vspace{-1ex}
\label{fig:framework}
\end{figure*}

%% file: sec_experiment.tex
\section{Experiments}
\label{sec:experiment}

\subsection{Transfer Learning Datasets}
\input{table/dataset.tex}
We perform experiments on 8 image classification datasets which span many different domains including common objects~\citep{imagenet,caltech101}, fine-grained animals~\citep{pets}, indoor and outdoor scenes~\citep{sun397}, foods~\citep{food101}, traffic signs~\citep{gtsrb},
natural textures~\citep{dtd}, and human actions~\citep{ucf101}.
Table~\ref{tbl:datasets} shows the detailed information of each dataset.

\subsection{\name Results}

Table~\ref{tbl:main} shows the unsupervised classification results on 6 tasks.
\name substantially improves upon CLIP on all tasks with an average accuracy improvement of \textbf{+11.0}\% / \textbf{+8.5}\% for ViT-B / ViT-L.
We also experiment with a widely-adopted supervised learning approach: supervised pre-training on ImageNet (1k or 21k) followed by supervised finetuning on the downstream tasks.
We finetune the ImageNet model with the same number of epochs and the same data augmentation~\citep{randaug} as \name.
Compared to the label-intensive supervised learning,
unsupervised \name achieves comparable performance on some datasets (e.g., Food101 and UCF101).
\name still lags behind supervised learning on some datasets (e.g., GTSRB and DTD).
We hypothesis that CLIP's low accuracy is the major cause.
Since CLIP has not seen enough data from those niche domains during pre-training,
its pseudo-labels would contain a limited amount of useful information.
Despite this,
\name can narrow the performance gap between unsupervised and supervised learning by around 50\%.
Therefore, MUST is a low-cost solution for image classification that opens up vast opportunities for practical scenarios where images are abundant but labels are scarce.

\input{table/main_result.tex}
\input{table/fewshot}
\subsection{Comparison with Few-shot Adaptation}
In Table~\ref{tbl:fewshot}, we compare \name with two state-of-the-art few-shot adaptation methods~\citep{coop,cocoop} that adapt a pre-trained CLIP model on labeled ImageNet images. 
\name outperforms 16-shot adaptation by a large margin (+6.2\%), demonstrating its advantage as a low-cost method to improve classification performance using unlabeled data.

\subsection{Unsupervised Transfer across Domains}
\label{sec:warmup}
\input{table/transfer.tex}
Since the \texttt{[MSK]} token and linear decoder are trained from scratch,
the potential of \name may not be fully exploited for downstream tasks with limited number of unlabeled images.
To address this, we propose to warm-up the model by training it on a similar domain with more unlabeled images.
Specifically,
we first employ \name to finetune a CLIP ViT-B model on ImageNet for a single epoch,
and then continue finetuning the model on two different datasets with limited number of samples (Pets and Caltech101).
During warmup,
we freeze the linear classifier (i.e., CLIP's text embeddings) to anchor the normalized image embeddings in their original space for easier transfer.
As shown in Table~\ref{tbl:transfer},
warmup on ImageNet further improves the performance of \name on both datasets.

\subsection{Ablation Study}

\noindent\textbf{Effect of the proposed objectives.}
We study the effect of the three objectives: self-training, mask image modeling, and global-local feature alignment.
The results are shown in Table~\ref{tbl:ablation}.
Removing the alignment loss reduces the average accuracy by 1.5\%,
and removing MIM further decreases accuracy by 0.7\%.
The results demonstrate the importance to align the features of \texttt{[CLS]} and \texttt{[MSK}] tokens,
which regularizes the model to learn better global features for classification.


\noindent\textbf{Number of \texttt{[MSK]} tokens to align with.}
\input{table/ablation.tex}
\input{table/align}
Our global-local feature alignment loss aims to align the \texttt{[CLS]} token to all of the \texttt{[MSK]} tokens for an image (see equation~\ref{eqn:align}).
We relax the alignment strength by only using the 10 \texttt{[MSK]} tokens that are nearest to the \texttt{[CLS]} token in the embedding space.
As shown in Table~\ref{tbl:align},
using the nearest-10 \texttt{[MSK]} tokens leads to similar performance compared to using all \texttt{[MSK]} tokens,
which suggests that \name is robust to the strength of alignment.

\noindent\textbf{Effect of fairness regularization.}
\input{table/fairness.tex}
We propose the fairness regularization loss $\mathcal{L}_\mathrm{reg}$ to counteract the bias in pseudo-labels.
In Table~\ref{tbl:fairness},
we examine its effect on the self-training objective.
The regularization improves the performance of self-training on all datasets.
It is the least helpful on SUN397, which has a long-tailed class distribution in its training set~\citep{sun397}.


\vspace{-0.5ex}
\subsection{Qualitative Analysis}

\input{figure/gradcam.tex}

\noindent\textbf{\name pays attention to more informative regions.}  
In Figure~\ref{fig:gradcam}, we visualize the GradCAM~\citep{gradcam} heatmap for the self-attention from \texttt{[CLS]} to all patches in the last ViT layer.
We compare three models - CLIP, ST, \name - and make the following qualitative observations:
(1) \name generally relies on more patches to make predictions. E.g., the fully body of the dog instead of only the head in row 1 column 1; all four corners of the traffic sign in row 3 column 1.
(2) \name can better leverage the true correlation between image patches and class labels, instead of the spurious correlations used by CLIP and ST. E.g., the guitar is attended for the action ``playing guitar'' in row 4 column 1, instead of the person or the background. 

\noindent\textbf{A helpful MIM is not necessarily good at image recovery.}  
In Figure~\ref{fig:mim_example},
we show examples of recovered images from the MIM output on the validation set of different datasets.
On some datasets with abundant samples (e.g., ImageNet),
the model can learn to recover masked patches with high quality.
On other datasets with fewer number of samples (e.g., GTSRB),
the recovered images have lower quality with many artifacts.
Despite the low recovery quality,
the MIM objective still provides useful supervision that improves the image classification performance.

\input{figure/example.tex}

\vspace{-0.5ex}
\subsection{Robustness to Image Corruptions}
\input{table/imagenet_c.tex}
In Table~\ref{tbl:imagenet_c},
we study the robustness of \name to image corruptions.
We evaluate the performance of ImageNet-finetuned models on the ImageNet-C\citep{imagenet_c} benchmark,
which contains 15 corruption types with various severity.
\name is more robust than CLIP and ST, as suggested by a lower mean corruption error (mCE).
The mCE of \name is only slightly higher than a model that is first trained with self-supervised MAE~\citep{mae} followed by \textit{supervised} finetuning on ImageNet.

\vspace{-0.5ex}
\subsection{Transductive Adaptation to Distribution Shift}
\input{table/transductive}
\input{figure/imagenet_ood}
We further investigate the robustness of \name under natural distribution shifts.
We experiment with three out-of-distribution variants of ImageNet:
ImageNetV2~\citep{imagenetv2} with the same 1000 ImageNet classes,
ImageNet-Rendition~\citep{imagenet-r} and ImageNet-Adversarial~\citep{imagenet-a} where each contains a subset of 200 classes (see Figure~\ref{fig:imagenet} for example images).
First, we directly evaluate a ViT-B model finetuned on ImageNet using \name, which has 77.7\% accuracy on ImageNet validation set.
As shown in Table~\ref{tbl:transductive},
\name on ImageNet improves the classification accuracy on ImageNetV2, but decreases the performance on ImageNet-R and ImageNet-A.
A similar observation is reported in CLIP~\citep{clip},
where supervised adaptation to ImageNet reduces the model's average robustness to out-of-distribution datasets.

To address this,
we perform transductive transfer learning~\citep{zeroshot_benchmark,transductive} with \name.
Transductive learning assume access to unlabeled test images from the new distribution,
which \name uses to jointly update model parameters and infer image labels.
The results in Table~\ref{tbl:transductive} show that transductive \name substantially improves the model's accuracy to distribution shift.
Furthermore, the warmup on ImageNet (Section~\ref{sec:warmup}) before transductive learning leads to more improvement.

\vspace{-1ex}
\section{Limitations}
\label{sec:limitation}
\vspace{-1ex}
In this paper,
we show \name as a promising domain-specific label-free classification algorithm where a different model is trained for each task.
Although the adaptation process is efficient, 
the storage of additional model parameters could be a practical limitation if the number of tasks is large. There exists a simple way to address this concern:
gather unlabeled image from all the domains of interest,
and perform \name to learn a single model that can generalize to multiple domains.
However, this may sacrifice the model's performance on each individual domain.

Due to the use of CLIP pre-trained models,
\name inherits some of the social implications of CLIP that arose from using minimally-filtered data crawled on the open internet.
Additional analysis on the model is necessary before deploying it for real-world applications.

%% file: table/dataset.tex
\begin{table*}[!t]
\small
\centering
\setlength\tabcolsep{3pt}
\resizebox{\textwidth}{!}{%
    \begin{tabular}	{l l l l l }
    Dataset  & Train size & Test size & Metric & Description \\
    \midrule
    ImageNet~\citep{imagenet} & 1,281,167 & 50,000  & acc & 1000 categories of objects from WordNet\\
    SUN397~\citep{sun397} & 76,129 & 21,758& acc  & 397 categories of indoor and outdoor scenes \\
    Food101~\citep{food101} & 75,750 & 25,250& acc  & 101 categories of food dishes \\
    GTSRB~\citep{gtsrb} & 26,640 & 12,630& acc  & 43 categories of German traffic signs \\
    DTD~\citep{dtd} & 3,760 & 1,880 & acc  & 47 categories of describable textures \\
    UCF101~\citep{ucf101} & 9,537 & 3,783 & acc  & 101 categories of human action video frames \\
    Oxford Pets~\citep{pets} & 3,680 & 3,669 & mean per class & 37 categories of cats and dogs \\
    Caltech101~\citep{caltech101} & 3,060 &  6,085 & mean per class & 101 categories of objects and 1 background\\
    
	\bottomrule
	\end{tabular}
}
\caption
{
\small	
	Information on the image classification datasets used for transfer learning.
}
\vspace{-1ex}
\label{tbl:datasets}
\end{table*}

%% file: table/main_result.tex
\begin{table*}[!t]
\small
\centering	
\setlength\tabcolsep{3pt}
\resizebox{\textwidth}{!}{%
	\begin{tabular}	{l l | l l l l l l  | l }
	& & ImageNet& SUN397& Food101& GTSRB& DTD&UCF101&	
	 Average \\
	 \midrule
	 \multirow{3}{*}{ViT-B/16} & CLIP & 68.3 & 64.4 & 88.7 & 43.4 & 44.7 & 68.8 & 63.1 \\
	 & \name  &77.7 \textcolor{Green}{(+9.4)} & 73.1 \textcolor{Green}{(+8.7)} & 92.7 \textcolor{Green}{(+4.0)} & 65.5 \textcolor{Green}{(+22.1)} & 54.1 \textcolor{Green}{(+9.4)} & 81.1 \textcolor{Green}{(+12.3)} & 74.1  \textcolor{Green}{(+11.0)} \\
	 & Supervised  & \textcolor{lightgray}{81.8} & \textcolor{lightgray}{77.2} & \textcolor{lightgray}{90.3} & \textcolor{lightgray}{97.1} & \textcolor{lightgray}{78.6} & \textcolor{lightgray}{82.3} & \textcolor{lightgray}{84.5} \\
	 \midrule
	 \multirow{2}{*}{ViT-L/14} & CLIP & 75.5 & 67.4 & 92.9 & 50.6 & 55.4 & 77.0 & 69.8 \\
	 &\name  & 82.1 \textcolor{Green}{(+6.6)} & 75.6 \textcolor{Green}{(+8.0)}& 95.3 \textcolor{Green}{(+2.4)}& 68.7 \textcolor{Green}{(+18.1)}& 62.6 \textcolor{Green}{(+7.2)}& 85.7 \textcolor{Green}{(+8.7)}& 78.3 \textcolor{Green}{(+8.5)}\\
\midrule
ViT-L/16 & Supervised &\textcolor{lightgray}{83.4} & \textcolor{lightgray}{81.1}&\textcolor{lightgray}{93.1} &\textcolor{lightgray}{97.3} & \textcolor{lightgray}{79.7}& \textcolor{lightgray}{86.4}& \textcolor{lightgray}{86.8}\\
	\bottomrule
	\end{tabular}
}

\caption
{
\small	
	Image classification results on a variety of downstream tasks. \name significantly improves upon CLIP for unsupervised classification on all datasets.
	Results for CLIP are obtained using the publicly released models.
	Supervised denotes the label-intensive common practice of supervised pre-training followed by supervised finetuning on downstream tasks.
	We use the ImageNet-1k model from~\cite{deit} for supervised ViT-B and the ImageNet-21k model from~\cite{vit} for supervised ViT-L.
}
\label{tbl:main}
\end{table*}		

%% file: table/fewshot.tex
\begin{table*}[!t]

\centering	

	\begin{tabular}	{l | c | c c | c }
	
	& ~CLIP~ & ~CoCoOp &CoOp& MUST  \\
	Finetuning data & none & \multicolumn{2}{c|}{16 labeled images per class} & unlabeled images\\ 
	 \midrule
	 ImageNet Acc. &  68.3 & ~71.0 & 71.5 & 77.7 \\

	\bottomrule
	\end{tabular}

\caption
{
\small	
	Comparison with few-shot CLIP adaptation methods on ImageNet.
   All methods use the same CLIP-pretrained ViT-B/16 model as image encoder.
}
\label{tbl:fewshot}
\vspace{-2ex}
\end{table*}		 

%% file: table/transfer.tex
\begin{wraptable}{r}{7.8cm}
\vspace{-3ex}
\small
\setlength\tabcolsep{3pt}
\resizebox{7.8cm}{!}{%
\centering	
	\begin{tabular}	{l | l l } 
    & Pets & Caltech101 \\
    \midrule
    CLIP & 88.9 & 89.8 \\
    \name w/o ImageNet warmup & 93.1 \textcolor{Green}{(+4.2)} & 93.1 \textcolor{Green}{(+3.3)} \\
    \name w/ ImageNet warmup & 94.3 \textcolor{Green}{(+5.4)} & 93.7 \textcolor{Green}{(+3.9)} \\
	\bottomrule
	\end{tabular}
}
\caption
{
\small	
Warmup on ImageNet improves the performance of \name on other domains.
}
\vspace{-1ex}
\label{tbl:transfer}
\end{wraptable}		

%% file: table/ablation.tex
\begin{table*}[!t]
\small
\centering	

	\begin{tabular}	{l | c c c c c c | l }
	\name objectives & ImageNet& SUN397& Food101& GTSRB& DTD&UCF101&Average \\
	 \midrule
	 ST+MIM+Align &77.7  & 73.1 & 92.7  & 65.5  & 54.1  & 81.1 & 74.1  \\
	ST+MIM  &77.1  & 72.0 & 92.4  & 62.5  & 52.6  & 79.2 & 72.6 (\textcolor{BrickRed}{\textendash1.5})  \\
	ST  &76.5  & 71.4 & 92.4  & 59.8  & 52.4  & 79.0 & 71.9 (\textcolor{BrickRed}{\textendash2.2})  \\
	\bottomrule
	\end{tabular}
\caption
{
\small	
	Effect of the proposed training objectives on unsupervised image classification.
}
\vspace{-1ex}
\label{tbl:ablation}
\end{table*}		 

%% file: table/align.tex
\begin{table*}[!t]
\small
\centering	

	\begin{tabular}	{l | c c c c c c | c }
	Num. \texttt{[MSK]} for $\mathcal{L}_\mathrm{align}$ & ImageNet& SUN397& Food101& GTSRB& DTD&UCF101&Average \\
	 \midrule
	All  &77.7  & 73.1 & 92.7  & 65.5  & 54.1  & 81.1 & 74.1  \\
	Nearest-10 & 77.6  & 73.2 & 92.7 & 65.7  & 54.1  & 80.8 &  74.1 \\
	\bottomrule
	\end{tabular}

\caption
{
\small	
	Number of \texttt{[MSK]} tokens used to align with the \texttt{[CLS]} token.
}
\label{tbl:align}
\end{table*}		 

%% file: table/fairness.tex
\begin{table*}[!t]
\small
\centering	

	\begin{tabular}	{l | c c c c c c | l }
	& ImageNet& SUN397& Food101& GTSRB& DTD&UCF101&Average \\
	 \midrule
	 ST w/ $\mathcal{L}_\mathrm{reg}$ &76.5  & 71.4 & 92.4  & 59.8  & 52.4  & 79.0 & 71.9  \\
	ST w/o $\mathcal{L}_\mathrm{reg}$ &74.0  & 71.0 & 91.8 & 50.0  & 46.8  & 73.0 &  67.8 (\textcolor{BrickRed}{\textendash4.1}) \\
	\bottomrule
	\end{tabular}

\caption
{
\small	
	Effect of the fairness regularization loss on self-training performance.
}
\label{tbl:fairness}
\end{table*}		 

%% file: figure/gradcam.tex
\begin{figure*}[!t]
\centering
  \includegraphics[width=\textwidth]{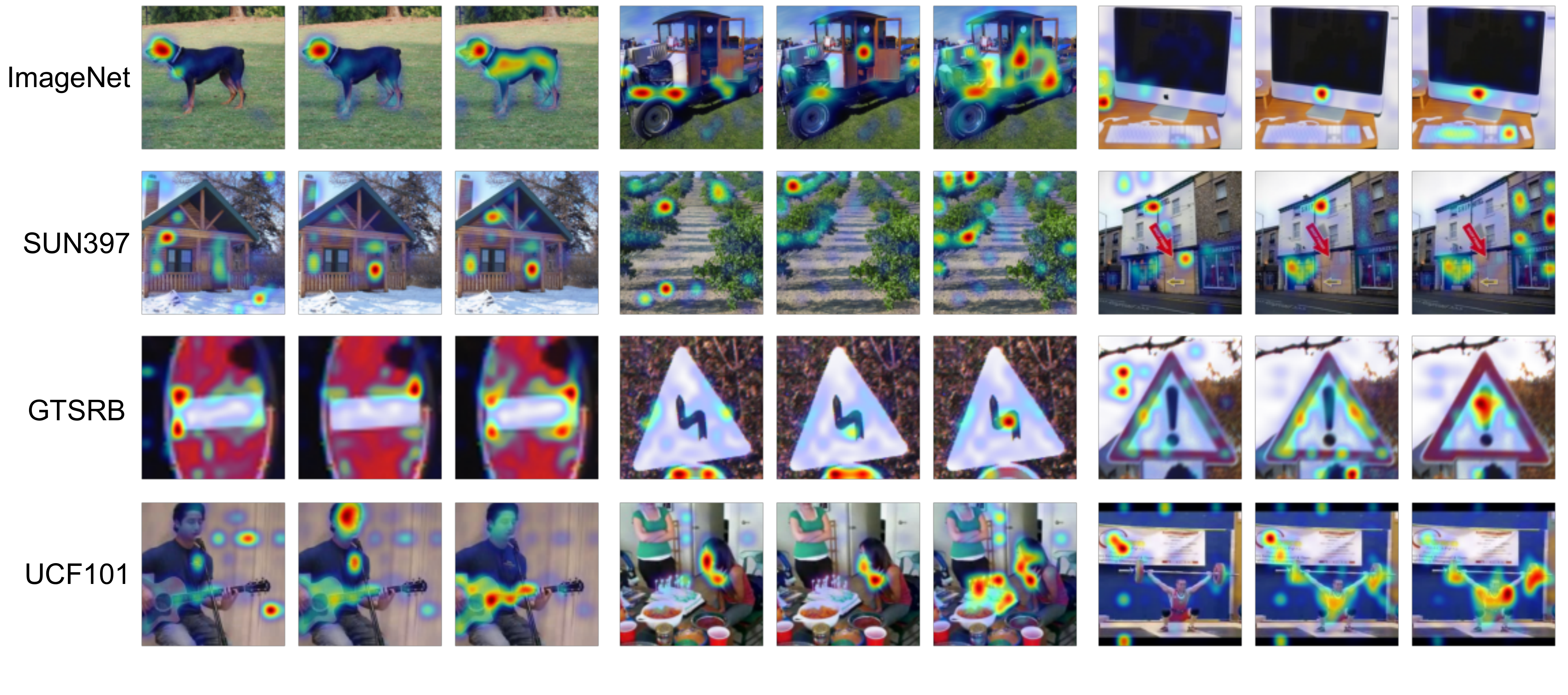}
\vspace{-4ex}
\caption{
\small 
GradCAM visualization on validation images from multiple datasets.
We visualize the self-attention from \texttt{[CLS]} to all patches in the last ViT layer.
For each triplet, we compare three different models: CLIP (left), ST (middle), \name (right).
We make two qualitative observations: (1) \name generally relies on more patches to make predictions; (2) \name can better leverage the true correlation between patches and class labels.
}
\vspace{-1ex}
\label{fig:gradcam}
\end{figure*}

%% file: figure/example.tex
\begin{figure*}[!t]
\centering
  \includegraphics[width=\textwidth]{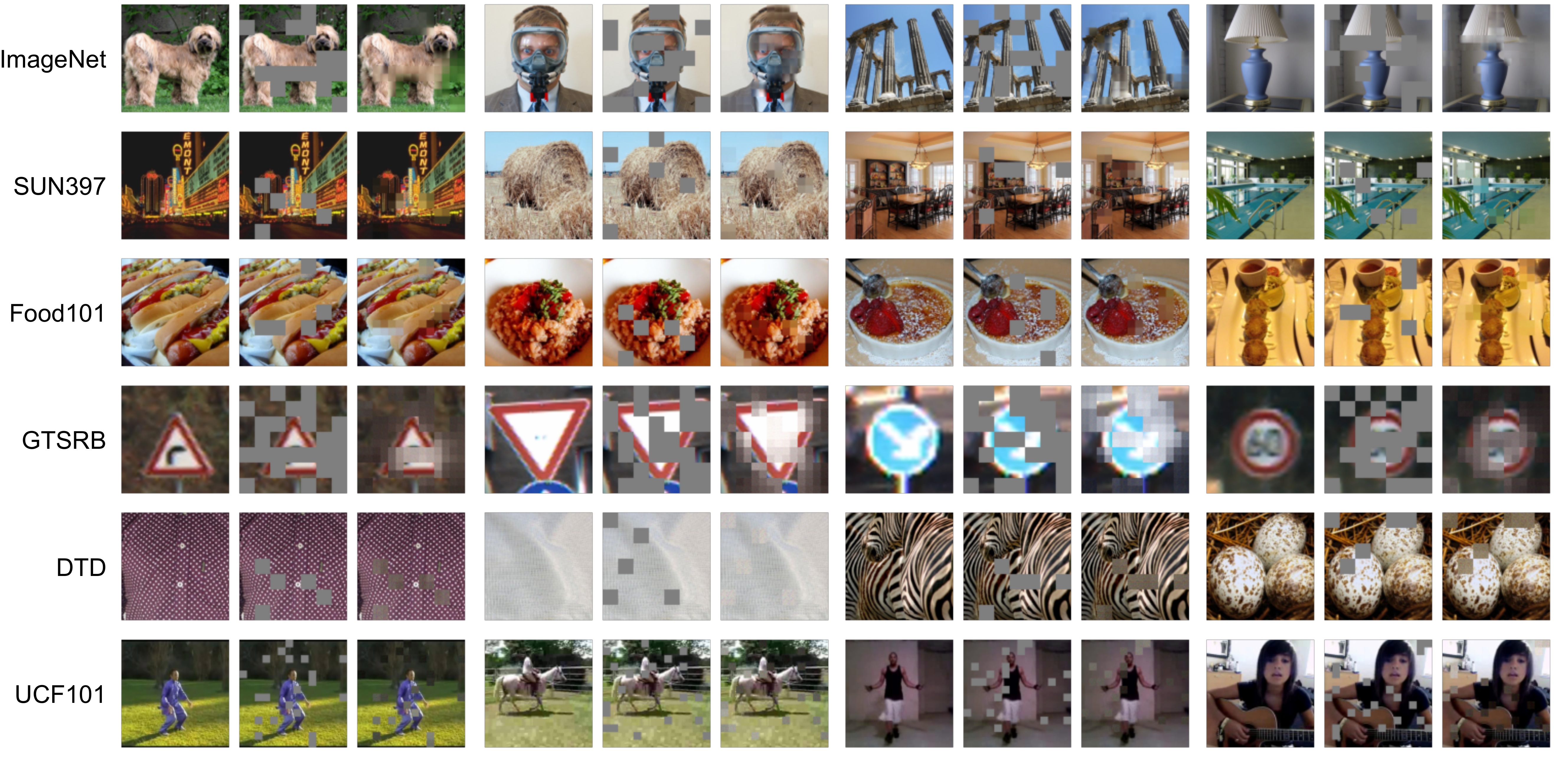}
\vspace{-4ex}
\caption{
\small
Examples of mask image modeling on validation images from multiple datasets.
For each triplet, we show the original image (left),
the masked image (middle), and the recovered image (right).
}
\vspace{-1ex}
\label{fig:mim_example}
\end{figure*}

%% file: table/imagenet_c.tex
\begin{table*}[!t]
\small
\centering	

	\begin{tabular}	{l | c c c c }
	ImageNet-C~\cite{imagenet_c} & CLIP & ST & \name & MAE+Supervised~\cite{mae} \\
	\midrule
	mCE $\downarrow$ & 70.7 & 55.2 & 53.4 & \textcolor{lightgray}{51.7} \\
	\bottomrule
	\end{tabular}

\caption
{
\small	
	Robustness evaluation on the ImageNet-Corruption dataset.
}
\vspace{-1ex}
\label{tbl:imagenet_c}
\end{table*}

%% file: table/transductive.tex
\begin{table*}[!t]
\small
\centering	
	\begin{tabular}	{l | c c c} 
    & ImageNetV2 & ImageNet-R  & ImageNet-A \\
    \midrule
    CLIP & 61.9 & 77.7 & 49.9 \\
    Inductive \name on ImageNet & 68.9 \textcolor{Green}{(+7.0)} & 68.7 \textcolor{BrickRed}{(\textendash
9.0)} & 41.6  \textcolor{BrickRed}{(\textendash
8.3)}\\
    Transductive \name w/o ImageNet warmup & 64.7 \textcolor{Green}{(+2.8)} & 87.3  \textcolor{Green}{(+9.6)}& 56.9 \textcolor{Green}{(+7.0)} \\
    Transductive \name w/ ImageNet warmup & 66.9 \textcolor{Green}{(+5.0)} & 87.6 \textcolor{Green}{(+9.9)} & 58.2 \textcolor{Green}{(+8.3)} \\
	\bottomrule
	\end{tabular}
\caption
{
\small	
Evaluation of \name under natural distribution shift from ImageNet.
}
\vspace{-2ex}
\label{tbl:transductive}
\end{table*}		

%% file: figure/imagenet_ood.tex
\begin{figure*}[!t]
\centering
  \includegraphics[width=\textwidth]{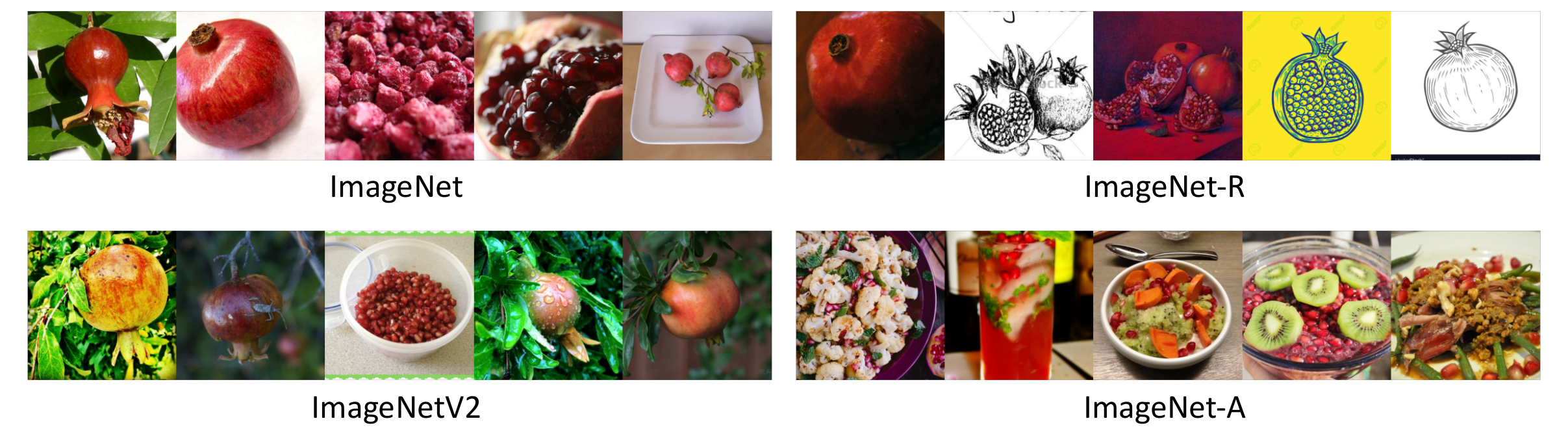}
\vspace{-4ex}
\caption{
\small 
Example images of \textit{pomegranate} from ImageNet and other datasets of natural distribution shifts.
}
\label{fig:imagenet}
\end{figure*}

%% file: sec_conclusion.tex
\vspace{-1ex}
\section{Conclusion}
\label{sec:conclusion}
\vspace{-1ex}

\input{table/parameters.tex}

The field of deep learning has been largely driven by simple and effective methods that can scale well.
\name is a simple method that can effectively adapt a pre-trained zero-shot classifier to downstream tasks in an unsupervised manner.
The core of \name is to simultaneously leverage two different sources of supervision signals obtained from unlabeled data:
task-specific supervision from pseudo-labels and task-agnostic self supervision from raw images.
We propose a simple alignment objective that bridges the two sources of supervision and enables the model to learn better features.

As a future work, \name can be naturally extended to other domains such as NLP,
where mask language modeling could act as the task-agnostic objective to improve self-training. We hope that this perspective could inspire future methods in unsupervised learning and zero-shot classification. 


%% file: table/parameters.tex
\begin{table*}[!t]
\small
\centering	
\setlength\tabcolsep{3pt}
\resizebox{\textwidth}{!}{%
	\begin{tabular}	{l | c c | c c| c c| c c| c c| c c| c| c }
	& \multicolumn{2}{c|}{ImageNet}& \multicolumn{2}{c|}{SUN397}& \multicolumn{2}{c|}{Food101}& \multicolumn{2}{c|}{GTSRB}& \multicolumn{2}{c|}{DTD}& \multicolumn{2}{c|}{UCF101}& Pets &
	Caltech101\\
	
	& B & L & B & L & B & L & B & L & B & L & B & L & B & B \\
	\midrule
	base lr & 2e-5 & 4e-5 & 1e-4 & 2e-4 & 2e-5 & 4e-5 & 2e-5 & 4e-5 & 2e-5 & 4e-5 & 1e-5 & 2e-5 & 1e-5 & 1e-5 \\
	epoch & 30 & 20 & 20 & 20 & 40 & 40 & 20 & 20 & 60 & 60 & 60 & 60  & 60 & 60 \\
	threshold $\tau$ & 0.7 & 0.5 & 0.4 & 0.3 & 0.7 & 0.7 & 0.3 & 0.2 & 0.4 & 0.5 & 0.5 & 0.6 & 0.7 & 0.7 \\
	mask ratio & 0.3 & 0.3 &0.1 & 0.1&0.1 & 0.1&0.5&0.3&0.1 & 0.1&0.1 & 0.1&0.1&0.1\\
	mask patch size & 32 & 56 & 32 & 28 & 32 & 56& 32 & 56& 32 & 56 & 16 & 14 & 32 & 32\\
	$\lambda_\mathrm{align}$& 0.2	& 0.5 & 1 & 0.5	&0.1 & 0.5	&0.5 & 0.5	&0.2 & 0.5	&	0.5&0.5&0.1&0.1\\
	$\lambda_\mathrm{reg}$& \multicolumn{2}{c|}{1} & \multicolumn{2}{c|}{0.2} & \multicolumn{2}{c|}{1} & \multicolumn{2}{c|}{1} & \multicolumn{2}{c|}{1} & \multicolumn{2}{c|}{1} & 1 & 1\\	
	EMA init. $\mu_0$ & \multicolumn{2}{c|}{0.999} & \multicolumn{2}{c|}{0.99} & \multicolumn{2}{c|}{0.999} & \multicolumn{2}{c|}{0.99} & \multicolumn{2}{c|}{0.99} & \multicolumn{2}{c|}{0.99} & 0.99 & 0.99\\
	EMA iters. $\mu_n$ & \multicolumn{2}{c|}{2000} & \multicolumn{2}{c|}{500} & \multicolumn{2}{c|}{2000} & \multicolumn{2}{c|}{500} & \multicolumn{2}{c|}{100} & \multicolumn{2}{c|}{500} & 100 & 100\\
	\bottomrule
	\end{tabular}
}
\caption
{
\small	
	Hyperparameters for \name on the downstream datasets.
	B: ViT-B/16; L: ViT-L/14.
}
\vspace{-1.5ex}
\label{tbl:hyper_parameters}
\end{table*}